# Multiframe Scene Flow with Piecewise Rigid Motion[*]

Vladislav Golyanik[1,2,3]   Kihwan Kim[1]   Robert Maier[1,4]   Matthias Nießner[4,5]   Didier Stricker[2,3]   Jan Kautz[1]

[1]NVIDIA   [2]University of Kaiserslautern   [3]DFKI   [4]Technical University of Munich   [5]Stanford University

## Abstract

*We introduce a novel multiframe scene flow approach that jointly optimizes the consistency of the patch appearances and their local rigid motions from RGB-D image sequences. In contrast to the competing methods, we take advantage of an oversegmentation of the reference frame and robust optimization techniques. We formulate scene flow recovery as a global non-linear least squares problem which is iteratively solved by a damped Gauss-Newton approach. As a result, we obtain a qualitatively new level of accuracy in RGB-D based scene flow estimation which can potentially run in real-time. Our method can handle challenging cases with rigid, piecewise rigid, articulated and moderate non-rigid motion, and does not rely on prior knowledge about the types of motions and deformations. Extensive experiments on synthetic and real data show that our method outperforms state-of-the-art.*

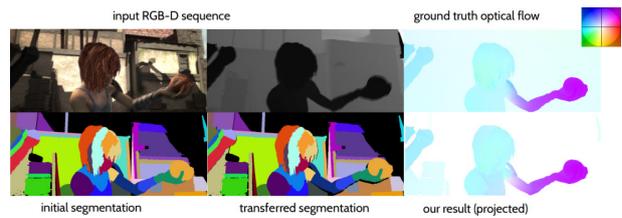

**Figure 1:** The proposed MSF approach accurately estimates scene flow between two or multiple frames. Taking a set of RGB-D frames as an input, it computes an oversegmentation of the reference frame and solves for coherent rigid segment transformations into the set of current frames. The recovered scene flow is highly accurate, with sharp motion boundaries. In the right column, the ground truth optical flow for the given input frames from the MPI SINTEL [4] data set is visualized using the Middlebury color scheme [3] (top), together with the projection of the scene flow onto the image plane estimated by our method.

## 1. Introduction

Scene flow refers to a 3D point velocity field of an observed scene, and has been a key element in various computer vision applications such as 3D reconstruction [29, 15], motion analysis and prediction [13, 24] as well as stereo matching [14, 8]. *Multiframe scene flow* (MSF) refers to a flow between a *reference* frame and every other frame of an image sequence (non-reference frames are referred as *current* frames). In the case of depth-augmented or RGB-D images, the input of a scene flow algorithm is a sequence of 2D images with corresponding depth maps. RGB-D based methods use the known depth measurements and inherently provide more accurate estimates compared to other classes.

At the same time, the accuracy of current RGB-D methods is still not sufficient for many application scenarios, even in the case of predominantly small rigid motion; it significantly deteriorates under *large* scene changes and multibody transformations (when multiple scene regions transform and deform independently). Moreover, the real-time requirement narrows down the choice of suitable optimization techniques and algorithmic solutions, which often sacrifices the accuracy in favour of the processing speed. Thus, an open question is increasing the accuracy (for small and large motions) without jeopardizing the speed. This motivated us to review all aspects of the modern RGB-D based scene flow estimation and find improvements for the bottleneck aspects.

We found *two major realms for improvements*. The first one is the piecewise rigid motion modelling combined with an oversegmentation of the scene. While the idea of piecewise rigidity was explored in the context of scene flow estimation in several ways before [16, 23, 11], we propose to combine a rigid parameterization with an oversegmentation of the reference frame, and jointly optimize for the movement of the segment pose pairs in a global manner (for multiple frames). In our model, we assume that object boundaries coincide with a subset of the segment boundaries. In other words, whenever there is an object boundary, a segment boundary must follow it; e converso, segment boundaries inside the objects can run arbitrarily. This is a reasonable assumption in practice, and reliable methods for such oversegmentation exist [1, 5]. Moreover, our method performs robustly even if the assumption about conciding boundaries is not entirely fulfilled, *i.e.*, it is robust against inaccuracies in oversegmentation. Another advantage is that segmentation updates result only in segment merging but not splitting. Requiring unconstrained continu-ous per-frame segmentation updates would result in significant increase of the solution space dimensionality, number of unknowns and the runtime. The proposed assumption allows to avoid those side effects.

---

[*]This work was partially supported by the BMBF project DYNAMICS (01IW15003).

The second realm considers the design of an energy functional as well as a choice of a robust optimization technique. Most current state-of-the art methods make use of total variation regularizers and employ variants of gradient descent or variational optimization for flow. In contrast, our energy functional is given in term of sums of squared residuals and optimized using linearization techniques. This strategy has proven to be efficient for multiple computer graphics and 3D reconstruction problems [29, 9]. In this paper, we show that *it is also highly effective for scene flow estimation*. A high-level overview of the proposed MSF appoach is shown in Fig. 1.

The remaining part of this paper is structured as follows: Related Work (Sec. 2), Proposed Approach (Sec. 3), Model Evaluation (Sec. 4) and Conclusion and Outlook (Sec. 5).

## 2. Related Work

Early methods for scene flow estimation required multiple consistent observations of optical flows [21, 28, 8, 24], and additionally solved for the unknown depths or disparities. Piecewise rigid scene flow exploits local rigidity of a scene as an additional constraint for scene flow estimation from stereo images [23]. Vogel *et al.* proposed a sliding window multiframe scene flow approach [22] which imposes consistency of planar patches across stereopairs in all frames.

Since the advent of affordable depth sensors such as Kinect, RGB-D based scene flow estimation became an active research area. In Semi-Rigid Scene Flow (SRSF), Quiroga *et al.* proposed to overparameterize point displacements by individual rigid body motions (twists) [16]. Assuming the rigid motion in the scene is predominant, SRSF estimates scene flow as a sum of a rigid component and a non-rigid residual. The SRSF energy consists of brightness constancy, depth variation and a sum of weighted total variations terms; it is decoupled for an alternating non-linear ROF-model optimization. SRSF was shown to handle small and simple motions well. A similar concept of rigid body motion parameterization was proposed in the SphereFlow [7] where the correspondences are searched within a spherical range. Jaimez *et al.* proposed a real-time variational Primal-Dual (PD-)Flow [10]. Its energy includes photometric and geometric consistency terms as well as a spatial total variation regularizer. A GPU implementation of a parallel primal-dual solver enabled real-time processing rates for RGB-D scene flow estimation. Several approaches jointly estimate segmentation and scene flow [11, 19]. Motion Cooperation (MC-)Flow [11] relies on a linear subspace model, *i.e.*, the velocity of every point is represented as a sum of estimated per-segment rigid motions. In MC-Flow, the scene segmentation is initialized with K-means on the depth channel and the energy is estimated in an alternating manner for motion fields and smooth rigid segmentation. Sun *et al.* rely on depth layering intrinsic to the depth measurements for estimation of scene segmentation and flow [19]. Both latter methods are computationally expensive.

While most approaches rely on the assumption of small motion, only a few approaches explicitly address the scenario with large displacements [27]. In contrast to the previous methods, we assign individual rigid body transformations to the segments retrieved on the depth channel, and use a unified non-linear least-squares optimization framework (it can potentially run in real-time on a GPU). Instead of a 2D depth reprojection error, we optimize for coordinate differences in 3D space (see Sec. 3). Moreover, none of the depth-augmented methods performs global optimization for a subsequence of frames. The global optimization allows us to estimate larger motions and contributes to the overall accuracy of the method. Another distinguishing property of our MSF approach is kernel lifting. While previous methods use robust kernels (*e.g.*, Tukey biweight), we employ kernel lifting, *i.e.*, kernel augmentation with fidelity weights. Our method generates qualitatively more accurate results compared to SRSR, PD-Flow as well as several optical flow methods [26, 20] (after scene flow projection).

RGB-D based scene flow is also used in non-rigid reconstruction [15, 9]. In DynamicFusion [15], depth measurements are warped to a static canonical view by a sparse displacement field and eventually integrated into a volumetric representation. The methods rely on the object texture and feature extraction, can handle moderate motion and support loop-closure but accumulate drift under large deformations. RGB-D scene flow can also be computed by non-rigid point set registration approaches such as *gmmreg* [12]. Point set registration methods recover 3D displacement fields between two or several point clouds. These methods constitute a separate class since differences in appearance and transforming displacement fields between the states can be large, and the consideration is not restricted to surfaces. The idea of scene segmentation has recently found its way in the optical flow estimation. Sevilla-Lara *et al.* showed that accurate object segmentation in 2D can significantly improve the accuracy of the optical flow estimation [17]. Similar to our case, their method requires an accurate object segmentation which follows the object boundaries. Since the depth channel is available in our case, we use a more accurate Felzenszwalb's segmentation algorithm [5] on the depth data.

## 3. Proposed Approach

Our method takes a sequence of N RGB-D image pairs as an input. We denote by $\mathcal{C}_i \in \mathcal{C}$ and $\mathcal{D}_i \in \mathcal{D}$ correspond-ing color and depth images respectively. We assume that all $\mathcal{C}_i$ and $\mathcal{D}_i$ are synchronized temporally as well as spatially (*i.e.*, registered). We work with the corresponding intensity images which are denoted by $\mathcal{I}_i$. The objective of MSF is reconstruction of a 3D displacement field $\rho(\mathbf{x}^t)$ of points $\mathbf{x}$ visible in the *reference* frame $\mathrm{N}_{\mathrm{ref}}$ into all remaining *current* $\mathrm{N} - 1$ frames of the observed scene. Thus, MSF relates 3D positions of every 3D point $\mathbf{P} = (P_x, P_y, P_z)$ in all observed frames as

$$\mathbf{P}^t = \mathbf{P}_{\mathrm{ref}} + \rho(\mathbf{x}^t). \quad (1)$$

We can handle the case when changes in a scene observed from $\mathrm{N}_{\mathrm{ref}}$ to every other frame are significant and, consequently, $\rho(\mathbf{x}^t)$ might contain large displacements (compared to frame-to-frame cases with small scene changes). A scene may consist of multiple independently moving rigid as well as non-rigid parts. The scene flow estimation in this case is an inherently ill-posed problem as multiple $\rho(\mathbf{x}^t)$ may lead to the same observed scene states.

In our model, every visible 3D Point $\mathbf{P}$ is projected onto the image plane with the projection operator $\pi : \mathbb{R}^3 \to \mathbb{R}^2$:

$$\mathbf{p}(x,y) = \pi(\mathbf{P}) = \left( f_x \frac{P_x}{P_z} + c_x, f_y \frac{P_y}{P_z} + c_y \right)^{\mathsf{T}}, \quad (2)$$

with $\{f_x, f_y\}$ the focal lengths of the camera and the principal point $(c_x, c_y)^{\mathsf{T}}$. The inverse projection operator $\pi^{-1} : \mathbb{R}^2 \times \mathbb{R} \to \mathbb{R}^3$ maps a 2D image point to a 3D scene point along the preimage given a depth value $z$ as follows:

$$\pi^{-1}(\mathbf{p}(x,y), z) = \left( z \frac{x - c_x}{f_x}, z \frac{y - c_y}{f_y}, z \right)^{\mathsf{T}}. \quad (3)$$

Furthermore, we assume scene transformations to be locally rigid, *i.e.*, the whole scene can be split into K segments $\mathcal{S}_k$, $k \in \{1, \ldots, \mathrm{K}\}$, and every point moves in the segment $\mathcal{S}_k$. By varying the segment granularity, different piecewise rigid and locally non-rigid motions can be accounted for. Movement of every segment $\mathcal{S}_k$ is given by its frame $l$ to frame $m$ rotation $\mathbf{R}_k^{l,m}$ and translation $\mathbf{t}_k^{l,m}$. We denote by $\mathbf{T}_k^{l,m} = (\mathbf{R}_k^{l,m}, \mathbf{t}_k^{l,m})$ the pose of the $k$-th segment. $\mathbf{T}_k^{l,m}$ has in total 6 DOF, *i.e.*, 3 DOF for $\mathbf{R}_k^{l,m}$ and 3 DOF for $\mathbf{t}_k^{l,m}$. $\mathbf{R}_k^{l,m}$ is parameterized through the angle-axis representation. In the angle-axis representation, a rotation is encoded by a vector $\boldsymbol{\alpha} = (\alpha_x, \alpha_y, \alpha_z)$. The direction of $\boldsymbol{\alpha}$ indicates the axis of rotation $\boldsymbol{\alpha}_n \in \mathfrak{so}(3)$ (it is obtained by normalization of $\boldsymbol{\alpha}$) — an element of Lie algebra — and the length of $\boldsymbol{\alpha}$ indicates the angle of rotation $\theta$ around $\boldsymbol{\alpha}_n$ according to the right-hand rule. To rotate a segment, we convert $\boldsymbol{\alpha}$ to the corresponding rotation matrix $\mathbf{R}$ using a corollary of the Rodrigues' rotation formula leading to the following exponential map:

$$\mathbf{R} = \exp(\theta \, \mathbf{K}) = \mathbf{I} + \sin\theta \, \mathbf{K} + (1 - \cos\theta) \mathbf{K}^2, \quad (4)$$

where $\mathbf{K} = \begin{pmatrix} 0 & -\alpha_z & \alpha_y \\ \alpha_z & 0 & \alpha_x \\ -\alpha_y & -\alpha_x & 0 \end{pmatrix} \in \mathfrak{so}(3)$ is given by a skew-symmetric cross-product matrix of $\boldsymbol{\alpha}_n$.

To recover scene flow $\rho(\mathbf{x}^t)$, we jointly solve for all segment poses $\mathbf{T}^{l,m}$ from every frame to every other frame in the input RGB-D sequence. Once the poses are recovered, MSF is estimated by considering per-point correspondences of the segments throughout the sequence. Our MSF approach is based on energy functional minimization. The energy $\mathfrak{E}$ in the two-frame case consists of four terms:

$$\begin{aligned}\mathfrak{E}(\mathbf{T}, \mathbf{w}) = {} & \alpha \, \mathfrak{E}_{\mathrm{data}}(\mathbf{T}) + \beta \, \mathfrak{E}_{\mathrm{pICP}}(\mathbf{T}) + \\ & + \gamma \, \mathfrak{E}_{\mathrm{l.\ reg}}(\mathbf{T}, \mathbf{w}) + \eta \, \mathfrak{E}_{\mathrm{r.\ opt.}}(\mathbf{w}).\end{aligned} \quad (5)$$

In Eq. (5), $\mathbf{T}_k \in \mathbf{T}$ are rigid transformations (rotation and translation) for every segment $k \in \{1, \ldots, K\}$ from the reference frame to the single current frame, $w_{j,h} \in \mathbf{w}$ is a set of lifting weights for the segment pairs optimized coherently. In the following, we describe each of the energy terms from Eq. (5) in detail. Prior to the energy-based optimization, we perform Gaussian smoothing of $\mathcal{I}_i$ to reduce the influence of noise and outliers when computing gradients (see Sec. 3.2 for optimization details).

**Data term.** The data term accounts for the brightness constancy of 3D points observed in multiple views $\mathcal{I}_i$. That is, the same 3D point $\mathbf{P}$ must cause the same brightness value in both views (similarly, the reprojection error of the brightness values associated with the same 3D point shall be minimized):

$$\begin{aligned}&\mathfrak{E}_{\mathrm{data}}(\mathbf{T}) = \\ &= \sum_k \sum_{\mathbf{p} \in \Omega_k} \left\| (\mathcal{I}_1(\mathbf{p}) - \mathcal{I}_2(\pi(g(\mathbf{T}_k, \pi^{-1}(\mathbf{p}, z_{\mathbf{p}})))))^2 \right\|_\epsilon,\end{aligned} \quad (6)$$

where $\Omega_k$ is a set of points in the segment $\mathcal{S}_k$, $g(\cdot, \cdot)$ is the rigid point transformation operator, $z_{\mathbf{p}} = \mathcal{D}_i(\mathbf{p})$ are known and $\|\cdot\|_\epsilon$ stands for the Huber loss defined as:

$$\left\| a^2 \right\|_\epsilon = \begin{cases} \frac{1}{2} a^2, & \text{for } |a| \leq \epsilon \\ \epsilon(|a| - \frac{1}{2}\epsilon), & \text{otherwise,} \end{cases} \quad (7)$$

with a non-negative scalar threshold $\epsilon$. The term in Eq. (6) performs dense image alignment of the current and reference intensity frames and contributes to the recovery of the relative segment transformations $\mathbf{T}_k$. The Huber loss works as an $\ell_2$ loss for the smallest values of the brightness residuals. If higher than $\epsilon$, the differences are not squared which is equivalent to an $\ell_1$ loss. As a result, the influence of significant non-Gaussian distributed outliers is reduced. The form of the Huber loss in Eq. (7) allows, de facto, to use an $\ell_1$ norm in the non-linear least squares framework.

**Projective ICP term.** The second data term in our target energy is the point-to-plane registration or projective Iterative Closest Point (ICP) term. In contrast to the intensity-based data term — which minimizes reprojected intensity values in the image space — the projective ICP minimizes Euclidean distances of corresponding points projected onto

the normals $\mathbf{n_p}$ of the reference image $N_{\text{ref}}$ directly in 3D:

$$\mathfrak{E}_{\text{pICP}}(\mathbf{T}) = \sum_k \sum_{\mathbf{p} \in \Omega_k} \left\| ((g(\mathbf{T}_k, \mathbf{p}) - \mathbf{p}^{\text{corr}}) \cdot \mathbf{n_p})^2 \right\|_\epsilon. \quad (8)$$

Every iteration of the projective ICP consists of two alternating steps: first, while segment poses are fixed, the correspondences $\mathbf{p}^{\text{corr}}$ for every point $\mathbf{p}$ are updated and, second, given point correspondences, the new segment poses $\mathbf{T}_k$ are computed. The optimum is achieved when the difference between two points is small and orthogonal to the normal $\mathbf{n_p}$ of the reference. The normals $\mathbf{n_p}$ are computed on the reference frame. For every point $\mathbf{p}$, the normal $\mathbf{n_p}$ is obtained as a cross product of the central differences $d_x$ and $d_y$ in the $x$ and $y$ directions respectively:

$$\mathbf{n_p} = d_x \times d_y. \quad (9)$$

The projective ICP term is brightness invariant and operates purely on the 3D points computed from the depth map. Thus, spatial diversity of a scene facilitates registrations of a higher accuracy.

**Lifted segment pose regularizer.** The data terms $\mathfrak{E}_{\text{data}}(\mathbf{T})$ and $\mathfrak{E}_{\text{pICP}}(\mathbf{T})$ alone are not sufficient to accurately align piecewise rigid scenes, especially when an accurate object-background segmentation of the reference frame might not be available. Without an explicit regularization, relatively small segments would be influenced by clustered outliers, noise, missing point correspondences and occlusions, among other disturbing effects. Thus, we propose the following term:

$$\mathfrak{E}_{\text{l. reg}}(\mathbf{T}, \mathbf{w}) = \sum_{w_{j,h}: \Psi[j,h]=1} \left\| (w_{j,h}^2 (\mathbf{T}_j - \mathbf{T}_h))^2 \right\|_2, \quad (10)$$

where $j$ and $h$ define a pair of segments with an imposed coherent movement and $\|\cdot\|_2$ is an $\ell_2$ norm. The segment pose regularizer favors coherent transformation of the neighboring segments.

The pairs of the segments which are demanded to move coherently are determined based on the segment vicinities. We maintain a segment adjacency matrix, *i.e.*, a sparse $K \times K$ matrix which contains ones if segments $j$ and $h$ move coherently, and zeroes otherwise. The number of segments can be large, and we store only non-zero elements of $\Psi$. We parameterize $\Psi = \psi_{j,h}$ with the number of adjacent segments $n_\psi$ for each segment. For every $\mathcal{S}_k$, a corresponding row of $\Psi$ contains $n_\psi$ segments with the closest centroids. For every pose pair, Eq. (10) defines a residual. Altogether, there are $\sum \psi_{j,h}$ residuals in the segment pose regularizer.

For the sake of robustness against outliers and disturbing effects mentioned above, we opt for lifting of the pose regularizer. Kernel lifting was shown to outperform robust costs (such as iterative reweighting, Triggs correction, *etc.*) concerning avoiding local minima [25]. The idea of lifting consists in augmenting the energy with confidence weights.

Thus, we introduce $w_{j,h}$ which in our formulation account for the strength of the segment connections and allow to continuously adjust segment coherencies. Recall that the scene might exhibit multiple independent motions and deformations, and $w_{j,h}$ allow to weaken or break the connections if two segments move independently. Kernel lifting works in combination with the next, robust weight optimizer term.

**Robust weight optimizer.** The lifted segment pose regularizer term alone is not sufficient to influence segment pair connectivities. The robust weight optimizer serves as a term preventing highly coherent segment pairs from weakening their weights:

$$\mathfrak{E}_{\text{r. opt.}}(\mathbf{w}) = \sum_{\forall w_{j,h}: \Psi[j,h]=1} \left\| (1 - w_{j,h}^2)^2 \right\|_2. \quad (11)$$

The operator introducing $\mathbf{w}$ into the energy is often referred to as a lifting function. We choose the lifting function

$$\mathscr{F}(\cdot, \mathbf{w}) = \sum (w_i^2(\cdot) + (1 - w_i^2)) \quad (12)$$

as it was shown to simplify the overall energy landscape well and also proven to be the most robust among other lifting kernel choices in our experiments. If optimally balanced, the segment pose regularizer and the robust weight optimizer can maintain strong connections by keeping $w_{j,h}$ high, even if the partial energy of the segment pose regularizer remains high.

### 3.1. Multiframe formulation

We generalize the energy proposed in Eq. (5) for the case of multiple frames. Consider a temporal window of N frames; we perform pairwise alignment of all frame pairs with individual data, projective ICP and segment pose regularizer terms but common weights $w_{j,h}$. This allows to further constrain the problem and enhance the accuracy (reduce multiframe registration drift) by adding more observations and performing the joint minimization. Let $i_{\text{ref}}$ and $i_{\text{target}}$ denote indexes of the first and the last frame of the temporal window and let Z be the set of pairwise frame combinations. In total, there are $^N C_2$ frame pairs, *i.e.*, a number of 2-combinations in a set of N frames. Moreover, let $\zeta \in Z$ be a one-dimensional variable indexing frame-to-frame transformations from frame $l$ to frame $m$, *i.e.*, $\mathbf{T}^\zeta$ is a shortcut for $\mathbf{T}^{l,m} \in Z$. Thus, the energy functional for the multiframe case reads

$$\begin{aligned}\mathfrak{E}(\mathbf{T}^1, \mathbf{T}^2, \ldots, \mathbf{T}^{|Z|}, \mathbf{w}) &= \sum_{\zeta \in Z} \alpha_\zeta \, \mathfrak{E}_{\text{data}}(\mathbf{T}^\zeta) + \\ &+ \sum_{\zeta \in Z} \beta_\zeta \, \mathfrak{E}_{\text{pICP}}(\mathbf{T}^\zeta) + \gamma_\zeta \sum_{\zeta \in Z} \mathfrak{E}_{\text{l.reg.}}(\mathbf{T}^\zeta, \mathbf{w}) + \\ &+ \eta \, \mathfrak{E}_{\text{r.opt.}}(\mathbf{w}) + \sum_{\zeta=3}^{|Z|} \lambda_\zeta \, \mathfrak{E}_{\text{c.}}(\mathbf{T}^\zeta).\end{aligned} \quad (13)$$

The global energy in Eq. (13) contains multiple data and projective ICP terms, lifted segment pose regularizer and robust weight optimizer terms (*cf.* the two-frame energy in Eq. (5)) as well as an additional pose concatenation term.

**Multiframe pose concatenation term.** Since we also optimize for poses between several frames with large displacements, the accumulated changes can be intractable for direct optimization. Additionally to the terms appearing in the energy functional in the two-frame case, we add a pose concatenation term which regularizes pose transformations to non-adjacent frames:

$$\mathfrak{E}_{\text{c.}}(\mathbf{T}^{l,m}) = \sum_k \left\| (\mathbf{T}_k^{l,m} - \mathbf{T}_k^{m-1,m} \cdot \ldots \cdot \mathbf{T}_k^{l,l+1})^2 \right\|_2, \quad (14)$$

with · denoting the pose concatenation operator. Thus, the pose concatenation term enforces the transformations between non-adjacent frames to be close to the concatenation of transformations which sequentially lead from frame $l$ to frame $m$.

### 3.2. Energy optimization

Our target energy consists of several terms, and every term contains a sum of squared residuals. Due to the perspective 3D to 2D and inverse 2D to 3D projections, the objective is non-linear and minimization of Eq. (13) is a non-linear least squares problem. We minimize it with a Gauss-Newton method. In total, there are M residuals:

$$\text{M} = {}^{\text{N}}C_2 \left( n_\mathcal{C} + n_\mathcal{D} \right) + ({}^{\text{N}}C_2 + 1) n_{\text{pp}} + n_{\text{c}}, \quad (15)$$

where $n_\mathcal{C}$ is the number of residuals in the two-frame data term ($\leq$ number of pixels in an image), $n_\mathcal{D}$ is the number of residuals in the two-frame projective ICP term (maximum the number of non-zero depth measurements); both $n_\mathcal{C}$ and $n_\mathcal{D}$ are appearing ${}^{\text{N}}C_2$ times. $n_{\text{pp}} = {}^{\text{N}}C_2 \sum \psi_{k,l}$ is the number of pose pairs (number of non-zero elements in the segment adjacency matrix) for every frame-to-frame combination. Additionally, there are $n_{\text{pp}}$ residuals in the robust regularizer (there are $n_{\text{pp}}$ weights in total). Finally, $n_{\text{c}} = \text{K}(\text{N} - 2)$ is the number of residuals in the pose concatenation term, *i.e.*, a number of non-adjacent frame combinations for given N per segment (if N = 2, then $n_{\text{c}} = 0$).

In the following, x is a shorthand symbol for the set of unknowns in the target energy $\mathfrak{E}$. We denote the scaled residuals by the compact notation $f_r(\text{x})$ and stack them into a single multivariate vector-valued function $\mathbf{F}(\text{x}) : \mathbb{R}^{\text{K}\,{}^{\text{N}}C_2 + n_{\text{pp}}} \to \mathbb{R}^{\text{M}}$:

$$\mathbf{F}(\text{x}) = [f_1(\text{x}), f_2(\text{x}), \ldots, f_r(\text{x})]^\mathsf{T}. \quad (16)$$

The total number of parameters is composed of K poses for each two-frame combination out of N and $n_{\text{pp}}$ weights. The objective function $\mathfrak{E}(x)$ can now be written in the new symbols as

$$\mathfrak{E}(x) = \|\mathbf{F}(\text{x})\|_2^2. \quad (17)$$

We aim at an optimal parameter set x' minimizing $\mathfrak{E}(x)$:

$$\text{x}' = \arg\min_{\text{x}} \|\mathbf{F}(\text{x})\|_2^2. \quad (18)$$

As the problem in Eq. (18) is non-linear, we iteratively linearize the objective around the current solution $\text{x}_t$ and find an update through minimization of the linear objective. The first-order Taylor expansion of (18) leads to

$$\mathbf{F}(\text{x} + \Delta \text{x}) \approx \mathbf{F}(\text{x}) + \mathbf{J}(\text{x})\Delta \text{x}, \quad (19)$$

where $\mathbf{J}(\text{x})_{\text{M} \times (\text{K}\,{}^{\text{N}}C_2 + n_{\text{pp}})}$ is the Jacobian of $\mathbf{F}(\text{x})$ at point x. Consequently, we define an objective for $\Delta \text{x}$:

$$\min_{\Delta \text{x}} \|\mathbf{J}(\text{x})\Delta \text{x} + \mathbf{F}(\text{x})\|^2. \quad (20)$$

Problem (20) is convex, and the minimum is achieved when

$$\mathbf{J}(\text{x})\Delta \text{x} = -\mathbf{F}(\text{x}). \quad (21)$$

Since (20) is overconstrained, Eq. (21) has a solution in the least-squares sense. By projecting $\mathbf{F}(\text{x})$ into the column space of $\mathbf{J}(\text{x})$, we obtain the corresponding normal equations

$$\mathbf{J}(x)^\mathsf{T} \mathbf{J}(\text{x}) \Delta \text{x} = -\mathbf{J}(\text{x})^\mathsf{T} \mathbf{F}(\text{x}) \quad (22)$$

which has a unique solution. For an enhanced convergence, we introduce a Tikhonov-type regularizer resulting in the Levenberg-Marquardt (LM) method:

$$\left(\mathbf{J}(x)^\mathsf{T}\mathbf{J}(\text{x}) + \lambda \operatorname{diag}(\mathbf{J}(x)^\mathsf{T}\mathbf{J}(\text{x}))\right) \Delta \text{x} = -\mathbf{J}(\text{x})^\mathsf{T} \mathbf{F}(\text{x}), \quad (23)$$

where $\lambda$ is a damping parameter. Beginning with a large $\lambda$ value, LM starts as a Gradient Descent method and changes over to the Gauss-Newton in the vicinity of the local minimum. To solve the system of linear equations (23), we apply Conjugate Gradient on the Normal Equations (CGNE). CGNE is a method from the Krylov subspace class. The main idea of CGNE consists in casting a solution of a linear system as a minimization of a quadratic form, and performing an update step in the direction accounting for the space stretching. CGNE takes advantage from the sparse matrix structure and is suitable for large systems. Though $\mathbf{J}(\text{x})^\mathsf{T}\mathbf{J}(\text{x})$ can be substantially large depending on the parameter number, $\mathbf{J}(\text{x})$ is highly sparse, as every individual residual depends only on few parameters from x. Thus, it neither stores nor computes $\mathbf{J}(\text{x})$ and $\mathbf{J}(\text{x})^\mathsf{T}\mathbf{J}(\text{x})$ explicitly. While optimizing $\mathfrak{E}$, we allow for non-monotonic steps. In the classic implementation, LM reverses to the previous state if a current update leads to an increased energy. Allowing suboptimal steps might result in the surpassing of local minima in the long term.

After computing the update $\Delta \text{x}$ in each iteration, we iteratively update the current solution as $\text{x}_{t+1} = \text{x}_t + \Delta \text{x}$. If the difference between several consecutive energy values continuously falls under the $\epsilon$ value, the optimization is considered as converged and the lowest energy value corresponding to the optimal x is returned. Once the energy is minimized, we compute correspondences between points of the reference and the current frames by 3D point reprojection. The scene flow $\rho$ is recovered as displacements between corresponding points. As a side effect, our approach can perform segmentation from motion. By grouping the recovered $\mathbf{T}_k$, we are able to determine independent rigid motions and deformations throughout the scene.

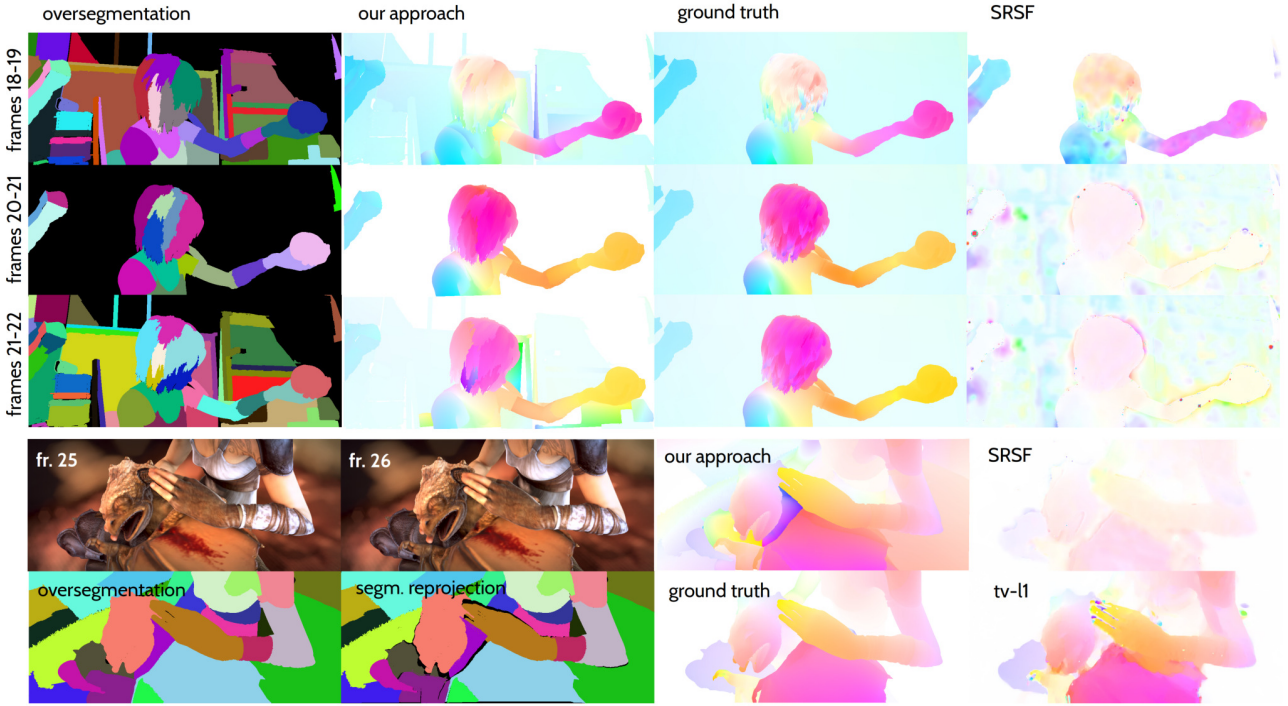

**Figure 2:** Experimental results on the SINTEL *alley1* (top three rows) and *bandage1* (bottom two rows) sequences. *alley1*, from the left to the right: oversegmentation for each reference frame, results of the proposed approach, ground truth optical flow and results of SRSF [16]; *bandage1*, on the left: the input frames, the initial oversegmentation and its re-projection into the current frame; on the right: results of our approach, SMSR [16], ground truth and the tv-l1 [26]. Due to optimally chosen segment sizes, our approach recovers an accurate scene flow compared to SRSF and tv-l1.

### 3.3. Energy initialization and settings

To alleviate the influence of noise during computation of image gradients in the brightness consistency term, we perform Gaussian smoothing of $\mathcal{I}_i$. We do not filter the depth data since the number of residuals in $\mathfrak{E}$ exceeds the number of parameters (segment poses and pose combination weights) by several orders of magnitude.

We initialize segments with graph-based segmentation on the depth values [5]. At the beginning, when no motion segmentation is available, the segment pairs are initialized based on the vicinity of the centroids (and so is the adjacency matrix $\Psi$). First, the center coordinates of the segments $\mathcal{S}_k$ are computed. Then, for every segment (row in the adjacency matrix), ones are set for K nearest segments (corresponding rows). We initialize the poses by aligning all neighboring frame pairs using the two-frame energy formulation. Transformations between non-adjacent frames are initialized with respective pose concatenations. We normalize every term in $\mathfrak{E}$ w.r.t. the total number of residuals. Thus, we set $\frac{\alpha^\varsigma}{M_{\alpha^\varsigma}}$, $\frac{\beta^\varsigma}{M_{\beta^\varsigma}}$, *etc.* as the term weights, where $M_{\alpha^\varsigma}$, $M_{\beta^\varsigma}$, *etc.* are the respective numbers of residuals.

## 4. Model Evaluation

In this section, we describe the experimental evaluation of the proposed technique. All experiments are performed under Ubuntu 16.04 on a system with 32 GB RAM and Intel Core i7-6700K CPU running at 4GHz. We implement MSF as a standalone framework in C++. As a non-linear least squares solver, we use *ceres* [2]. For the sake of accuracy and speed, we opt for automatic differentiation of cost functions. Since every flow vector is parameterized by 6 DOF, every corresponding residual depending on rigid transformation is a 6-vector. After every successful solver step, we update point correspondences for the projective ICP term. The algorithm converges in average after $15-20$ iterations.

### 4.1. Experiments on synthetic data

We use two synthetic data sets for evaluation — MPI SINTEL [4] and v-(irtual)KITTI [6]. SINTEL data set represents several sequences of synthetically rendered images of an animated movie with an additional depth chan-nel (the resolution is $1024 \times 436$ pixel). The imaging process (motion blur, defocus) and the atmospheric effects are accurately simulated so that the images look naturalistic. vKITTI represents a synthetic data set of a frequently encountered urbane driving scenes. We compare MSF with SRSF [16] as well as tv-l1 optical flow method of Zach *et al*. [26]. Comparison with PD-Flow is performed on the real data (Sec. 4.2), as it is hard-coded for Kinect recordings. Unfortunately, the source code for MC-Flow [11] is not publicly available.

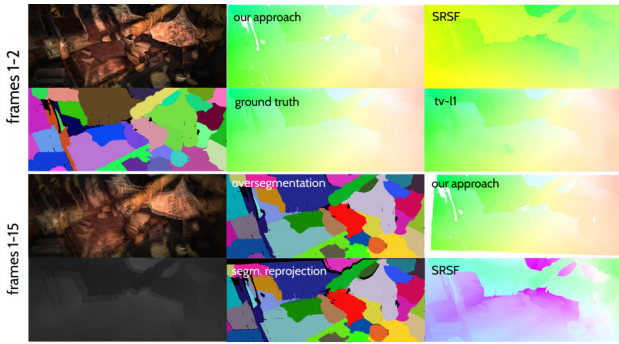

**Figure 3:** Experimental results on a static scene observed by a moving camera from the SINTEL data set (*sleeping2*). First column: input frames, an oversegmentation, large displacement frames and corresponding depth maps. Both SRSF[16] and our approach can incorporate prior knowledge of the scene rigidity while solving for the 3D flow. Both for the case of small (one frame) and large (14 frames) displacements, our approach outputs a very accurate result. We notice from the input images that in the case of large displacements, the camera preserves the trajectory; thus, direction of the flow does not change significantly in this scene. This effect is reflected in the projection of the scene flow estimated by our approach.

|          | *alley1*          | *bandage1*        | *sleeping2, rigid* |
|----------|-------------------|-------------------|--------------------|
| SRSF [16]| 2.46122/2.40833   | 2.47801/2.46389   | 1.13584            |
| MSF      | 0.740127          | 1.69865           | 0.307526           |

**Table 1:** Quantitative comparison between scene flow projections and the ground truth optical flow on the MPI SINTEL [3] training sequences using Average EPE errors (in pixel).

|          | 2 fr. | 3 fr. | 4 fr. | 2 fr. (r) | l. d. (r) |
|----------|-------|-------|-------|-----------|-----------|
| SRSF [16]| 274   | n.a.  | n.a.  | **87.5**  | 84.5      |
| MSF      | **49**| 221   | 541   | 90        | 254       |

**Table 2:** Average runtime comparison of the proposed Multiframe Scene Flow (MSF) and SRSF [16] for different configurations, in seconds. Legend: (r) for rigid scene; l.d. for large displacements (14 frames).

We set the minimum number of vertices per segment to $2 \cdot 10^3$ (if less, a segment is discarded) and the threshold to $0.5$ in the Felzenszwalb's algorithm [5]. The oversegmentation is used for parameterization of the scene flow by a set of rigid 6-DOF transformations. When the number and size of the segments are optimal, parametrization explains observed rigid motion and approximates non-rigid deformations well so that our results consistently outperform the competing methods. In this experiment, we have empirically determined the neccesary condition on the initial segmentation: *the segments boundaries must follow the object boundaries*. The more completely this condition is fulfilled, the more accurate can be the result. On the other side, if segmentation is not accurate, the accuracy decays insignificantly, up to a breakpoint. Fig. 2 shows representative results (*i.e.*, not biased) on the *alley1* and *bandage1* sequences. Projection of the scene flow and optical flow are visualised with the Middlebury color code [3].

Next, we compare MSF on the scene undergoing a purely rigid transformation — *sleeping2*. Similar to our approach, SRSF allows parameterizing the whole displacement field by a single rigid transformation, given prior knowledge of a scene. Fig. 3 shows results of the comparison for the case of small (one frame) and large (14 frames) differences. In the case of one frame difference, the ground truth is available. Our approach recovers a similar flow to the ground truth, whereas SRSF misses the direction by several degrees. In the case of large displacements, no ground truth is available, but the flow preserves the direction (this can be noticed from the images). MSF recovers the scene flow accurately also in this case, while the discrepancy of SRSF increases.

We evaluate accuracy of MSF quantitatively by comparing projections of the scene flow with the ground truth optical flow on the SINTEL data set. Table 1 reports the average End Point Error (EPE)[1] for SRSF [16] and our approach for several image sequences. The obtained metrics agree with the qualitative results. Table 2 lists average runtimes for SRSF [16] and MSF on the SINTEL data set, for different launch configurations. Additionally, we report EPE of our MSF method for *shaman2* (0.354061), *shaman3* (1.00192), *mountain1* (1.6898) and *bandage2* (1.0715) sequences.

With the example of the *alley1* sequence, we study the influence of the Huber loss on the scene flow result. Using the $\ell_1$ norm instead of the Huber loss (a Huber loss with a zero threshold) leads to the decrease in the runtime and accuracy by $5\%$ and $4\%$ respectively. Choosing the $\ell_2$ norm leads to an $\approx 40\%$ increase of EPE.

Evaluation on the vKITTI data set demonstrated the applicability of MSF for driving scenarios. Fig. 4-(bottom row) shows two consecutive frames from a driving scene as point clouds overlayed and warped to the current frame after scene flow estimation. Although the motion of the cars differs in direction and magnitude, both scene subflows are recovered accurately.

### 4.2. Experiments on real data

We also evaluate the proposed approach on several real data sets, *i.e.*, rigid multibody data set [18] recorded by a Kinect. We compare MSF with SRSF [16], PD-Flow [10] as well as tv-l1 optical flow [26] and Multi-Frame Subspace

---
[1]EPE is defined as $\|(u - u_{\text{GT}}), (v - v_{\text{GT}})\|$, where $(u, v)^\mathsf{T}$ is a projected flow vector and $(u_{\text{GT}}, v_{\text{GT}})^\mathsf{T}$ is a ground truth vector.

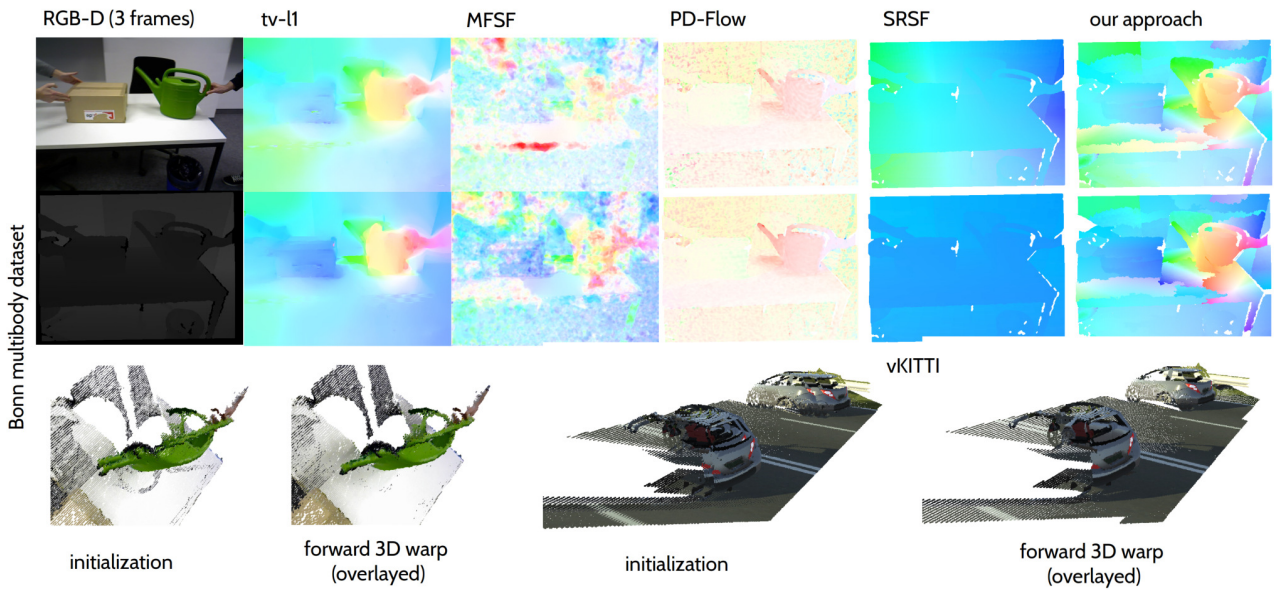

**Figure 4:** Results of several RGB-D scene flow and optical flow approaches on the Bonn multibody data set [18]. Top row: results for the frames 1-2; middle row: results for the frames 1-3; bottom row: initialized (not aligned) and warped (aligned) RGB-D frames visualized as point clouds (left, processed by MSF); an example of an urbane driving scene processing form the vKITTI data set [6] by MSF (right).

Flow (MFSF) [20]. Selected results are shown in Fig. 4. Since no ground truth is available for this real data set, several qualitative observations can be made. First, the watering pot moves towards the observer and the motion between three frames is rather small. MFSF, PD-Flow and SRSR generate more noisy output than tv-l1 and our approach. MSF preserves object boundaries, but some spurious boundaries are introduced which are more apparent than in the case of the SINTEL sequences. Obtaining accurate segmentation on noisy real world data requires additional pre-processing and can be readily done. Fig. 4-(bottom left) shows initialized and warped overlayed RGB-D frames visualized as point clouds.

### 4.3. Discussion

The experiments show advantages of the new scene flow formulation. We believe that the segmentation along with the requirement of sharp object boundaries in combination with suitable segment sizes reason the notable accuracy of MSF. In some cases, however, SLIC [1] or a simple regular tiled segmentation can be more suitable than Felzenszwalb though this is case-dependent. Another reason might be the projective point-to-plane ICP term which endows MSF with a property of an articulated point set registration algorithm. Finally, an improved lifted energy landscape is perhaps a part of the answer.

Yet, MSF has some limitations. If a scene does not provide sufficient cues for the data terms (brightness and varying depth values), segments which include such regions may be influenced by the moving parts in the vicinity and erroneously involved in the motion (cf. Fig. 2, our result on frames 21-22 — the segments near the girl's face get spuriously involved in the motion). The method also requires a balance between the lifted Laplacian pose regularizer and robust weight regularizer terms. If all weights are coming close to $1.0$ in the absolute value, $\eta$ needs to be *decreased*. Parameters should be set so that the weights $w_{ij}$ are distributed around $0.0$ and $1.0$ or $-1.0$ with a low variance. In all experiments, $\gamma$ and $\eta$ were fixed, while $\alpha$ and $\beta$ were set depending on the data properties.

### 5. Conclusion and Outlook

We propose a novel multiframe scene flow approach. Our method relies on an oversegmentation of the reference frame with sharp object boundaries. The underlying energy functional includes brightness constancy, projective ICP, robust Laplacian as well as chained pose regularizer terms, and each of them consists of sums of squared residuals. As a result, we obtain a robust method which performs accurately on multiple synthetic and real data sets and favourably compares to multiple RGB-D based scene flow and optical flow approaches including state-of-the-art. We believe that more improvements in RGB-D based scene flow estimation can be achieved within the proposed framework. In future work, we will focus on real-time performance aspects, since MSF is well parallelizable and can be implemented on a GPU. Moreover, we will evaluate the proposed model regarding the suitability for visual odometry and occlusion detection problems.